\documentclass[lettersize,journal]{IEEEtran}
\usepackage{amsmath,amsfonts}
\usepackage{algorithmic}
\usepackage{algorithm}
\usepackage{array}
\usepackage[caption=false,font=normalsize,labelfont=sf,textfont=sf]{subfig}
\usepackage{textcomp}
\usepackage{stfloats}
\usepackage{url}
\usepackage{verbatim}
\usepackage{graphicx}
\usepackage{cite}
\usepackage{wrapfig}
\usepackage{cleveref}
\usepackage{tabularray}
\usepackage{ulem}
\usepackage{booktabs}
\usepackage{color}
\usepackage{ulem} 
\begin{document}

\title{FaSDiff: Balancing Perception and Semantics in \\
Face Compression via Stable Diffusion Priors}

\author{Yimin Zhou, Yichong Xia, 
Bin Chen~\IEEEmembership{Member,~IEEE}, Mingyao Hong, Jiawei Li, \\ Zhi Wang~\IEEEmembership{Senior Member,~IEEE}, Yaowei Wang~\IEEEmembership{Member,~IEEE}
        % <-this % stops a space
\thanks{Y. Zhou, Y. Xia and Z. Wang are with Tsinghua Shenzhen International Graduate School, Tsinghua University, Shenzhen, Guangdong, 518055, China, Y. Xia also with Research Center of Artificial Intelligence, Pengcheng Laboratory, Shenzhen, Guangdong, 518055, China.(e-mail: zhou-ym24, xiayc23, wangzhi@mails.tsinghua.edu.cn)}% <-this % stops a space
\thanks{M. Hong is with Research Center of Artificial Intelligence, Pengcheng Laboratory, Shenzhen, Guangdong 518055, China. (e-mail: hongmy@pcl.ac.cn)}
\thanks{B. Chen is with Harbin Institute of Technology, Shenzhen, Guangdong, 518055, China, and also with Research Center of Artificial Intelligence, Pengcheng Laboratory, Shenzhen, Guangdong 518055, China.(e-mail: chenbin2021@hit.edu.cn.)(\textit{Corresponding author: Bin Chen.})}
\thanks{J. Li is with  Huawei Manufacturing, Shenzhen, Guangdong, 518055, China.(e-mail: li-jw15@tsinghua.org.cn) }
\thanks{Y. Wang is with Harbin Institute of Technology, Shenzhen, Guangdong, 518055, China, and also with Research Center of Artifcial Intelligence, Pengcheng Laboratory, Shenzhen, Guangdong 518055, China.(e-mail: wangyw@pcl.ac.cn)}}

\markboth{Journal of \LaTeX\ Class Files,~Vol.~14, No.~8, August~2021}%
{Shell \MakeLowercase{\textit{et al.}}: A Sample Article Using IEEEtran.cls for IEEE Journals}

\IEEEpubid{0000--0000/00\$00.00~\copyright~2021 IEEE}

\maketitle

\begin{abstract}

With the increasing deployment of facial image data across a wide range of applications, efficient compression tailored to facial semantics has become critical for both storage and transmission. While recent learning-based face image compression methods have achieved promising results, they often suffer from degraded reconstruction quality at low bit rates. Directly applying diffusion-based generative priors to this task leads to suboptimal performance in downstream machine vision tasks, primarily due to poor preservation of high-frequency details. In this work, we propose FaSDiff (\textbf{Fa}cial Image Compression with a \textbf{S}table \textbf{Diff}usion Prior), a novel diffusion-driven compression framework designed to enhance both visual fidelity and semantic consistency. FaSDiff incorporates a high-frequency-sensitive compressor to capture fine-grained details and generate robust visual prompts for guiding the diffusion model. To address low-frequency degradation, we further introduce a hybrid low-frequency enhancement module that disentangles and preserves semantic structures, enabling stable modulation of the diffusion prior during reconstruction. By jointly optimizing perceptual quality and semantic preservation, FaSDiff effectively balances human visual fidelity and machine vision accuracy. Extensive experiments demonstrate that FaSDiff outperforms state-of-the-art methods in both perceptual metrics and downstream task performance.
\end{abstract}

 
\section{Introduction}
\label{sec:intro}
With the increasing pursuit of convenience and privacy in daily life, facial images are being extensively utilized in applications such as identity verification, social networking, and virtual displays. Every day, billions of facial image data are captured, stored, and transmitted, necessitating efficient facial image compression techniques to support their storage and transmission. Efficient compression not only reduces the storage and bandwidth requirements for facial data, effectively minimizing operational costs, but also ensures high-quality reconstruction of facial images, thereby maintaining the accuracy of downstream tasks and enhancing user experience.

\begin{figure}[t]
    \centering
    \includegraphics[width=\linewidth]{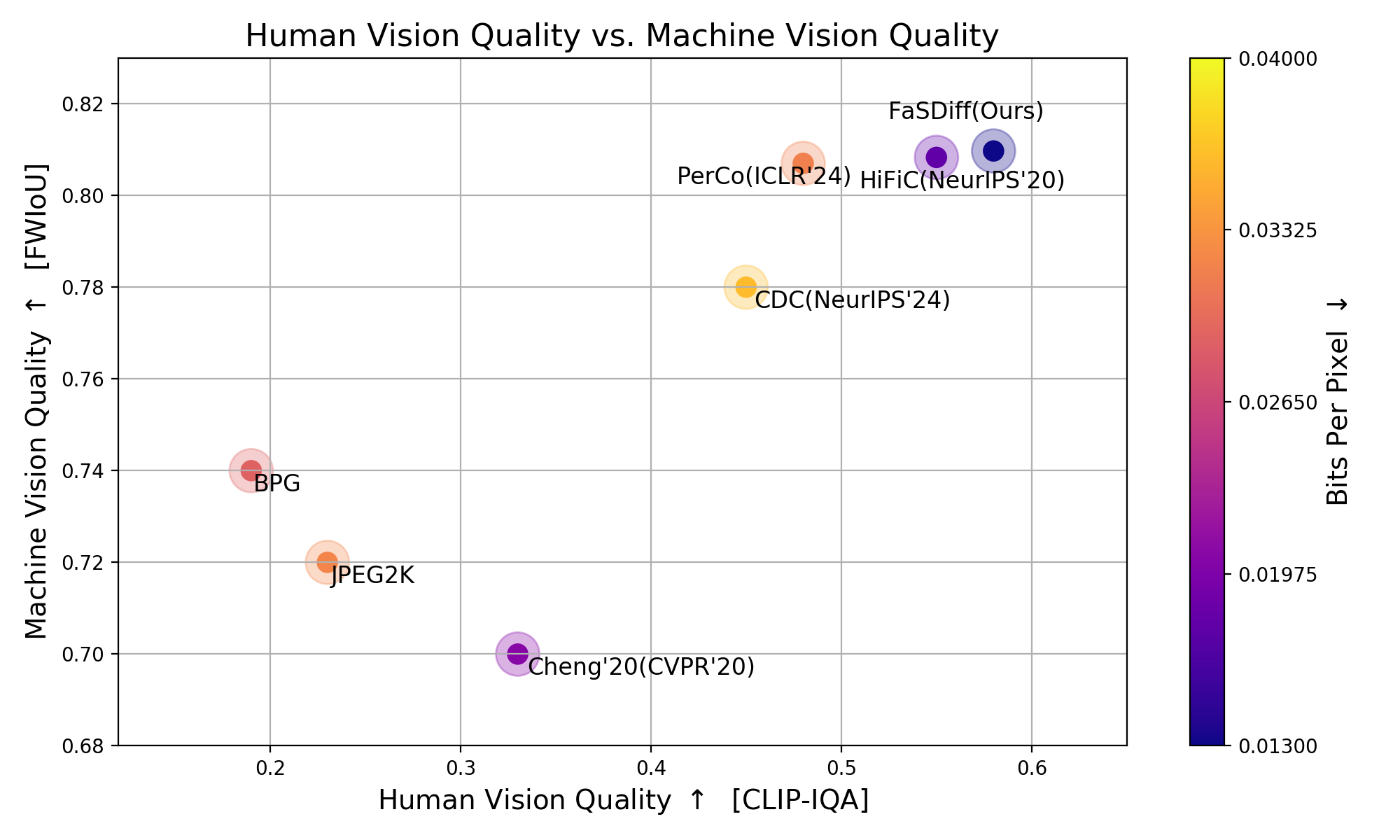}
    \caption{The trade-off between different compression methods for perceptual quality and downstream task performance. Proximity to the upper right corner indicates superior overall model performance. The color indicates the level of compression rate.} 
    \label{pic:first_fig}
\end{figure}
\begin{figure*}[t]
    \centering
    \includegraphics[width=1.0\textwidth]{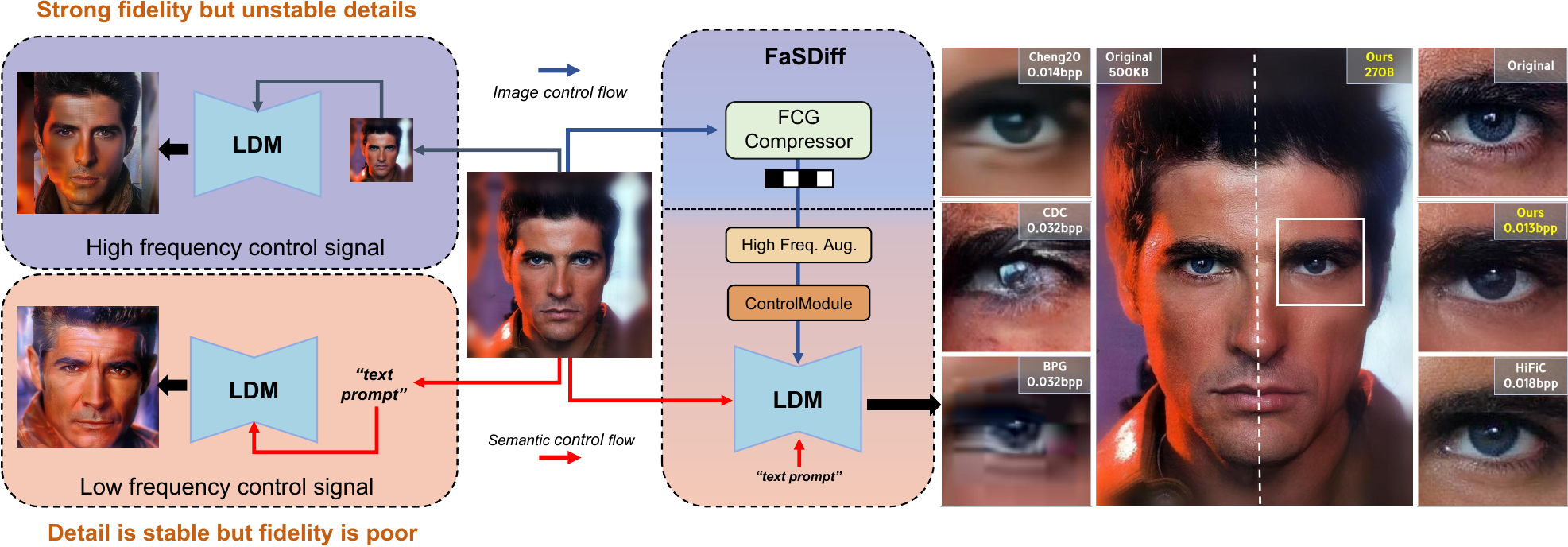}
    \caption{(Left): The basic T2I LDM generation mode struggles to produce controllable outputs with stable details while maintaining strong fidelity in the images. (Right): Overview of our proposed FaSDiff and qualitative comparison with mainstream solutions. FaSDiff employs a blend of low-frequency and high-frequency control, reconstructing intricate details at extremely low bit rates with perfect realism. } 
    % \vspace{-0.5cm}
    \label{pic:cvpr_pic1}
\end{figure*}
Compared to natural image compression tasks, facial image compression presents unique challenges. First, unlike the complex natural images that contain diverse high-frequency information, facial images possess distinct and prominent features with relatively uniform high-frequency content. Consequently, efficiently and intelligently allocating the bitstream to preserve critical facial information becomes a primary focus for facial image compression models. Secondly, the human visual system is exceptionally sensitive to facial details. Any compression-induced artifacts, blurring, or noise—which might be negligible in natural images—can be rapidly detected in facial images. Furthermore, a significant portion of stored and transmitted facial images is utilized in downstream tasks such as gender recognition and facial segmentation. This requirement necessitates that the reconstruction results from facial compression models are compatible with a variety of downstream tasks. Natural image compression methods do not typically account for these specific applications, often resulting in suboptimal performance when applied to facial image-related tasks
.

\IEEEpubidadjcol
In recent years, numerous studies have explored the use of deep neural networks for facial image compression. However, these approaches have yet to fully address the unique challenges associated with this task and still struggle to achieve extremely low compression ratios. Alternatively, other approaches \cite{wu2020general, mao2023scalable, yang2023facial} utilize generative adversarial networks (GANs) \cite{brock2018large, goodfellow2014generative} with generative priors to compensate for the details lost during compression, thereby enabling substantially lower bitrates. Nonetheless, due to inherent limitations of GAN models and issues related to training losses, these methods often exhibit severe image artifacts at low bitrates. As Text to Image (T2I) diffusion models emerge as powerful new generative frameworks \cite{saharia2022photorealistic,zhang2023adding}, there have been numerous attempts to apply these models in the field of compression \cite{careil2023towards,yang2024lossy}. However, when directly adapting them to facial compression tasks, existing approaches tend to overly rely on low-frequency information while neglecting high-frequency components. 
This reliance leads to difficulties in effectively balancing the trade-off between the perceptual quality of the generated images and the quality required for downstream tasks.
\begin{figure*}[t]
    \centering
    \includegraphics[width=0.9\textwidth]{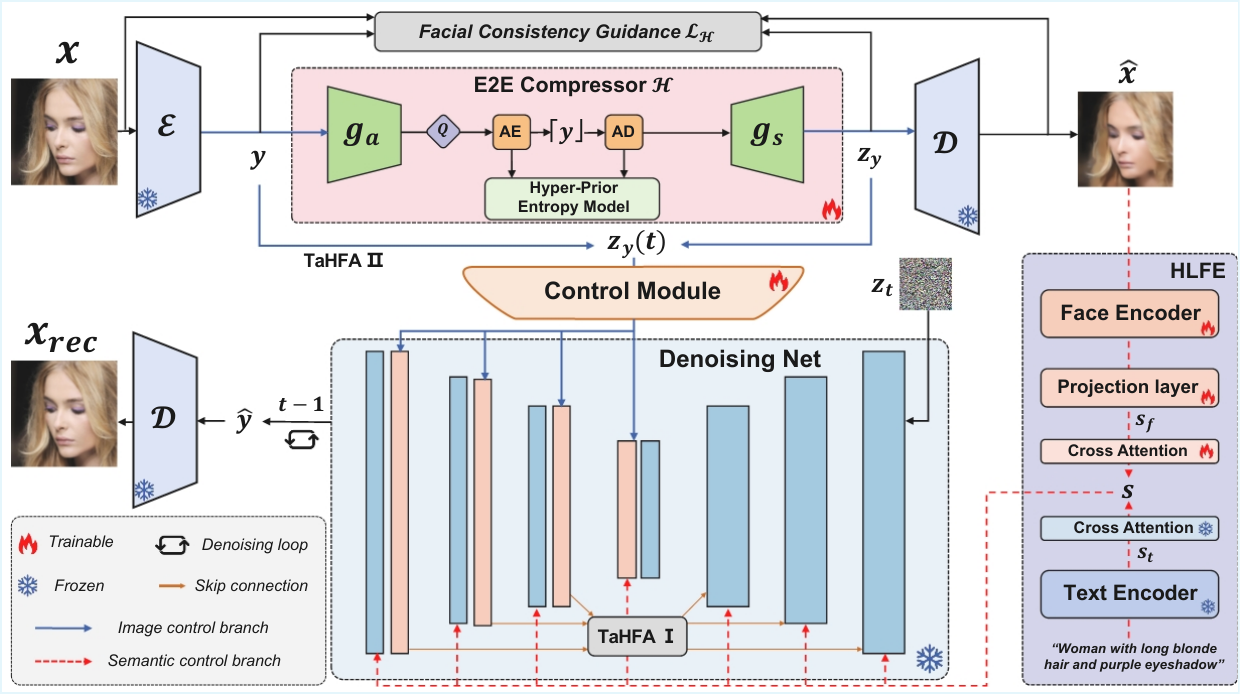}
    \caption{Illustration of the proposed  \textbf{Fa}cial Image Compression with a \textbf{S}table \textbf{Diff}usion prior (FaSDiff) framework. Initially, we extract $\boldsymbol{y}$ of the input image $\boldsymbol{x}$ through encoder $\mathcal{E}$. Subsequently, guided by facial consistency loss, we employ an end-to-end compressor to obtain the compressed high-frequency image control flow (solid blue line). Simultaneously, we decouple low-frequency facial semantics from the image prompts, generating a hybrid low-frequency semantic control flow (dashed red line). Ultimately, under the guidance of the cross-prompt interaction between high and low frequencies, the diffusion prior generates the final result $\boldsymbol{x}_{rec}$.} 
    \label{pic:pipeline}
\end{figure*}

To leverage the diffusion prior for generating high-quality images while avoiding the loss of consistency, we propose \textbf{Fa}cial Image Compression with a \textbf{S}table \textbf{Diff}usion prior (FaSDiff). Specifically, FaSDiff incorporates a Time-aware High-Frequency Augmentation (TaHFA) module, which enables the high-frequency-sensitive compressor to capture the details and textures of the latent image representations while preventing feature domain shifts through consistency loss. Subsequently, we reflected on the marginal effectiveness of CLIP embeddings in facial compression tasks. We introduced a hybrid low-frequency enhancement structure, decoupling strong semantic conditions from image prompts to stabilize the colors and details generated by denoising networks alongside textual prompts. As shown in \cref{pic:cvpr_pic1}, FaSDiff successfully preserves the advantages of the latent diffusion model at the perceptual level of the human and effectively mitigates the decline in visual task performance caused by the loss of semantic consistency, achieving an optimal balance between machine vision and human vision. Our comprehensive experimental results demonstrate the outstanding capability in facial image compression.

In short, our main contributions can be summarized as follows.

\begin{itemize}
    \item We introduce a novel facial image compression framework based on a foundational diffusion model: FaSDiff. FaSDiff can capture rich high-frequency signals with very few bits and leverage the priors of a pre-trained LDM to reconstruct images with perfect realism while maintaining facial feature consistency. 
    \item We integrate powerful hybrid semantic embeddings as additional prompts. Our FaSDiff enables the decoupling of advanced facial features from the image, enhancing the stability and semantic consistency of the diffusion model priors more effectively.
    \item FaSDiff has achieved the best trade-off between machine vision and human vision on facial image datasets, as depicted in \cref{pic:first_fig}. It demonstrates performance equivalent to that of the original images in machine while reaching the optimal performance in human vision.
\end{itemize}

\section{Related Work}
\label{sec:related_work}

%------------------------------------------------------------------------
%------------------------------------------------------------------------
\subsection{Facial Image Compression}

Facial image compression has been explored in early studies, with related region-based approaches reported in \cite {tropf2005region}, who proposed a novel region segmentation approach to identify facial components, thereby enhancing the accuracy of compression specifically for facial regions. Subsequently, studies such as \cite{elad2007low} and \cite{sujatha2016compression} employed various mathematical algorithms and statistical techniques to improve the effectiveness of facial image compression further.

With the advancement of deep learning and the integration of deep neural networks into the field of lossy image compression, \cite{wang2019scalable} introduced deep neural networks to facial image compression. Additionally, \cite{zhang2024principal} proposed the PCANet for image compression, which advances the field by leveraging principal component analysis within a neural network framework to optimize compression performance.

The advent of Generative Adversarial Networks (GANs) has enabled the generation of high-quality, high-fidelity, and high-resolution images. For facial images, generative models provide robust priors that significantly reduce the amount of information required for accurate reconstruction, making it possible to compress facial images for storage and transmission at an extremely low bit rate. \cite{wu2020general} introduced GANs into the image compression domain, achieving superior compression and reconstruction performance compared to existing methods.

Further exploration by subsequent studies, such as \cite{yang2023facial}, delved into generative adversarial models by applying GAN inversion techniques to facial image compression. This approach demonstrated scalability and achieved high-performance facial image compression at extremely low bitrates, marking a significant advancement in the field. Building on this, \cite{mao2023scalable} investigated the relationship between StyleGAN priors and specific facial features in greater depth, which enabled selective transmission and decoding based on downstream task requirements, thereby achieving exceptionally low bitrates while maintaining high reconstruction quality.
%-------------------------------------------------------------------------
\subsection{Compression with Diffusion Prior}

% Diffusion models \cite{sohl2015deep} have emerged as a powerful class of generative models, exhibiting remarkable performance in image synthesis and beyond. Rooted in the principles of stochastic differential equations, diffusion models, such as Denoising Diffusion Probabilistic Models (DDPM) introduced by \cite{ho2020denoising} operate by gradually adding Gaussian noise to data in a forward diffusion process and then learning to reverse this process to generate high-fidelity images. Unlike Generative Adversarial Networks (GANs), diffusion models are inherently stable to train and capable of capturing diverse data distributions without mode collapse, making them particularly suitable for tasks requiring high-quality and diverse image generation.

Leared Image Compression based on CNN architecture \cite{balle_end--end_2017, balle_variational_2018, cheng2020learned, he_elic_2022, jiang_mlic_2024, lifrequency,qin2025cassic,zeng2025mambaic} and GAN architecture \cite{hific, illm,jia2024generative,wu2025conditional} has been widely studied in recent years. With the advancement of generative models, diffusion models have shown great power in many tasks. In the field of image compression, numerous efforts have been made to integrate diffusion models. The diffusion model was first employed in \cite{yang2024lossy}. In such method, the image was first mapped onto a contextual latent variable which can be compressed into bitstreams, and then the compressed representation was utilized as a conditional guide for the denoising process, enabling the iterative generation of the reconstructed image. 

Following this, \cite{careil2023towards,relic2024lossy, li2024towards, xiadiffpc} attempted to incorporate the priors from pre-trained diffusion generative models, such as Stable Diffusion \cite{rombach2022high}, into image compression tasks. These methods adopt ControlNet or ControlNet-like paradigms, using compressed representations as conditional guidance to enable pre-trained diffusion models to refine and restore original images as much as possible while preserving the visual effects of the reconstructed images.By leveraging the robust generative priors of these pre-trained diffusion models, a greater proportion of bits allocated for storage and transmission can be assigned to low-frequency information, while high-frequency details are synthesized using the generative priors. This methodology makes great success and facilitates achieving extreme compression ratios without compromising image fidelity, resulting in higher-quality reconstructed images.

Another category of image compression methods based on diffusion priors was proposed by DiffC\cite{theis2022lossy}. These methods utilize the diffusion prior by directly predicting and restoring the noise introduced during the Gaussian noising process. Subsequently, PSC\cite{elata2024zero} proposed a posterior-based compression approach. After that, \cite{vonderfechtlossy} addressed the challenges of reverse-channel coding, successfully applying the DiffC algorithm to generative models in the Stable Diffusion series. Since they only predict noise without modifying the generation process, these methods can be directly applied to pretrained diffusion models without any additional training, thus boasting broader application scenarios. However, due to the gap between the assumption of quantization noise and the Gaussian assumption in diffusion generative models, such methods often only achieve suboptimal performance.
\section{Method}
\label{sec:method}

\subsection{Overall Framework}
The overall framework of FaSdiff is illustrated in \cref{pic:pipeline}, where the image to be encoded is denoted as $\boldsymbol{x}$, and the final decoded image is represented as $\boldsymbol{x}_{rec}$. We introduce an end-to-end compressor $\mathcal{H}(\cdot,\cdot,\cdot)$ that takes $\boldsymbol{y}$ and multi-level features $e_1,e_2$ from encoder $\mathcal{E}$, yielding an estimation $\boldsymbol{z}_{\boldsymbol{y}}=\mathcal{H}(\boldsymbol{y},e_1,e_2)$. The encodable quantized hidden representation $\lceil\boldsymbol{y}\rfloor$ of $\boldsymbol{z}_{\boldsymbol{y}}$ will be transmitted to the decoding end along with the textual description $\boldsymbol{s}_t$ of the input image $\boldsymbol{x}$.

%$\mathcal{H}$ predicts entropy parameters based on a hyperprior entropy model, estimating the probability density of the quantized latent representation of $\boldsymbol{y}$ and encoding it. We acquire the latent representation $\boldsymbol{y}$ of the image and intermediate layer features $e_1$ and $e_2$ through the pre-trained encoder $\mathcal{E}$. 

At the decoding end, $\boldsymbol{z}_{\boldsymbol{y}}$ is initially decoded by the pre-trained decoder $\mathcal{D}$ to obtain a preliminary estimation $\hat{\boldsymbol{x}}$, which is then input into a pre-trained facial feature extractor to acquire low-frequency semantic embeddings $\boldsymbol{s}_f$. Subsequently, $\boldsymbol{s}_f$ will be combined with $\boldsymbol{s}_t$ and fed into a modulation layer to obtain a fused low-frequency control signal $\boldsymbol{s}$. Next, $\boldsymbol{s}$, along with $\boldsymbol{z}_{\boldsymbol{y}}$, will modulate the features of the denoising model as guidance, resulting in the denoised output $\hat{\boldsymbol{y}}$. Finally, the decoded image $\boldsymbol{x}_{rec}$ will be obtained through the decoder $\mathcal{D}$ from $\hat{\boldsymbol{y}}$.

% These embeddings, along with the textual semantics $\boldsymbol{s}_t$, are fed into a modulation layer to obtain low-frequency control components $\boldsymbol{s}$. These low-frequency controls, along with the high-frequency control $\boldsymbol{z}_{\boldsymbol{y}}$, collectively serve as conditional terms to modulate features in the pre-trained LDM, resulting in the sampling result $\hat{\boldsymbol{y}}$, which, when decoded, yields the reconstructed image $\boldsymbol{x}_{rec} = \mathcal{D}(\hat{\boldsymbol{y}})$.

\begin{figure}[t]
    \centering
    \includegraphics[width=0.8\linewidth]{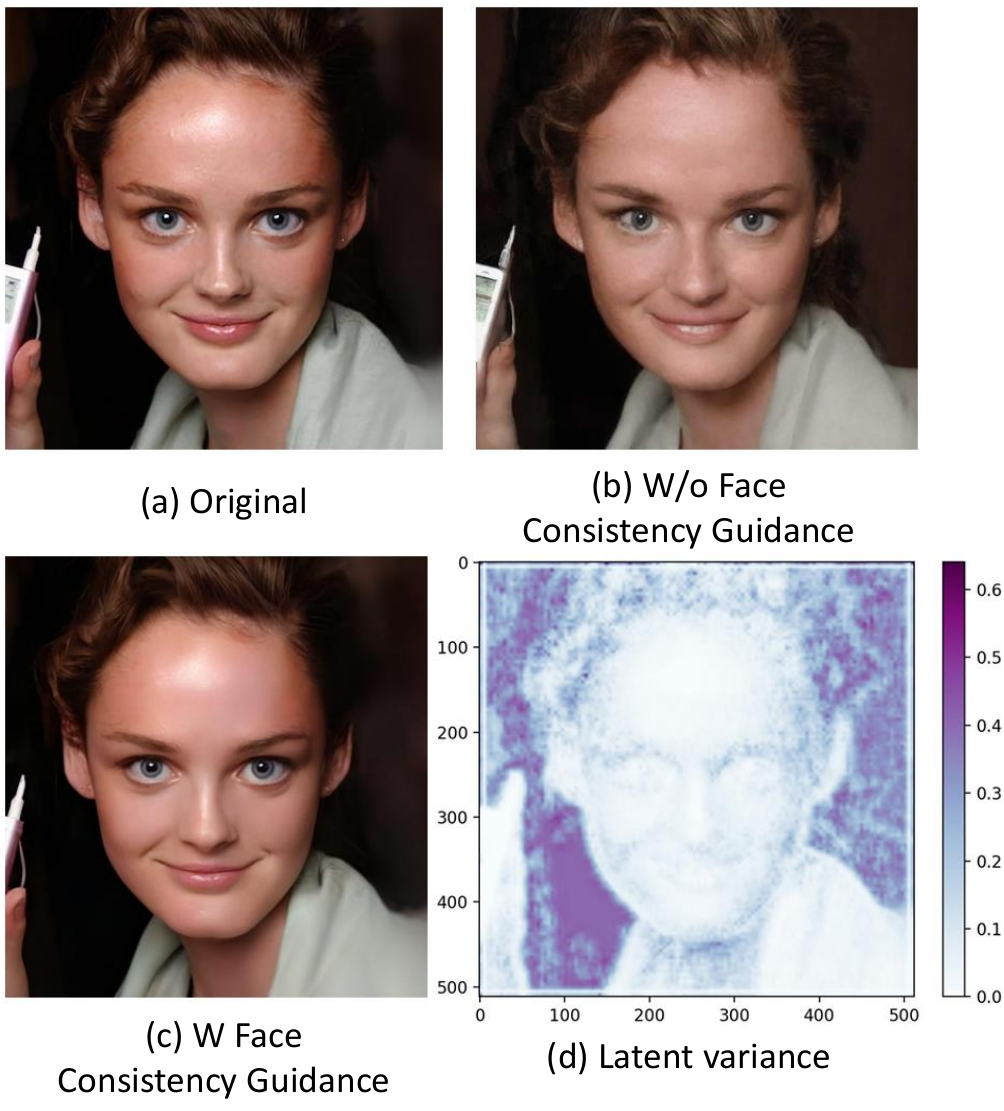}
    \caption{(a)-(c): An example using face consistency guidance. The face consistency loss makes the generated facial expressions more faithful to the original image. (d): Visualization results of the standardized variance $var(\boldsymbol{y})$.} 
    \label{pic:fig2_fcg}
\end{figure}
\subsection{Compressor Guidance for Facial Consistency}
\label{sec:FCG}
It is challenging to faithfully reconstruct images using diffusion priors. In low bit-rate scenarios, compression algorithms based on diffusion architectures often lose a significant amount of high-frequency image signals, greatly impacting the visual quality of reconstructed images. As shown in \cref{pic:fig2_fcg}(b), unconstrained diffusion baselines tend to generate highly realistic images but overlook details like the pose and expressions of the facial image. Therefore, we enhance the compressor to capture more high-frequency control signals through variance-weighted facial consistency loss and Time-aware high-frequency augmentation. At the decoding end, stable sampling of high-frequency details in images and fidelity of image embeddings is achieved through facial mixed semantic features.

As depicted in \cref{pic:pipeline}, we employ an end-to-end compressor that accepts multi-level features as input, encoding low-dimensional embeddings of $\boldsymbol{x}$. This approach ensures consistency between the decoding space and latent space while avoiding the loss of high-frequency information during encoding by $\mathcal{E}$. To guarantee that the decoding latent representation encompasses as many high-frequency signals as possible and ensures consistency in details such as facial features, we observe that the variance of $\boldsymbol{y}$ is detail-sensitive. As illustrated in \cref{pic:fig2_fcg}(d), regions concerning facial features and contours exhibit minimal variance, which is precisely the area of focus.

\begin{figure}[t]
    \centering
    \includegraphics[width=1\linewidth]{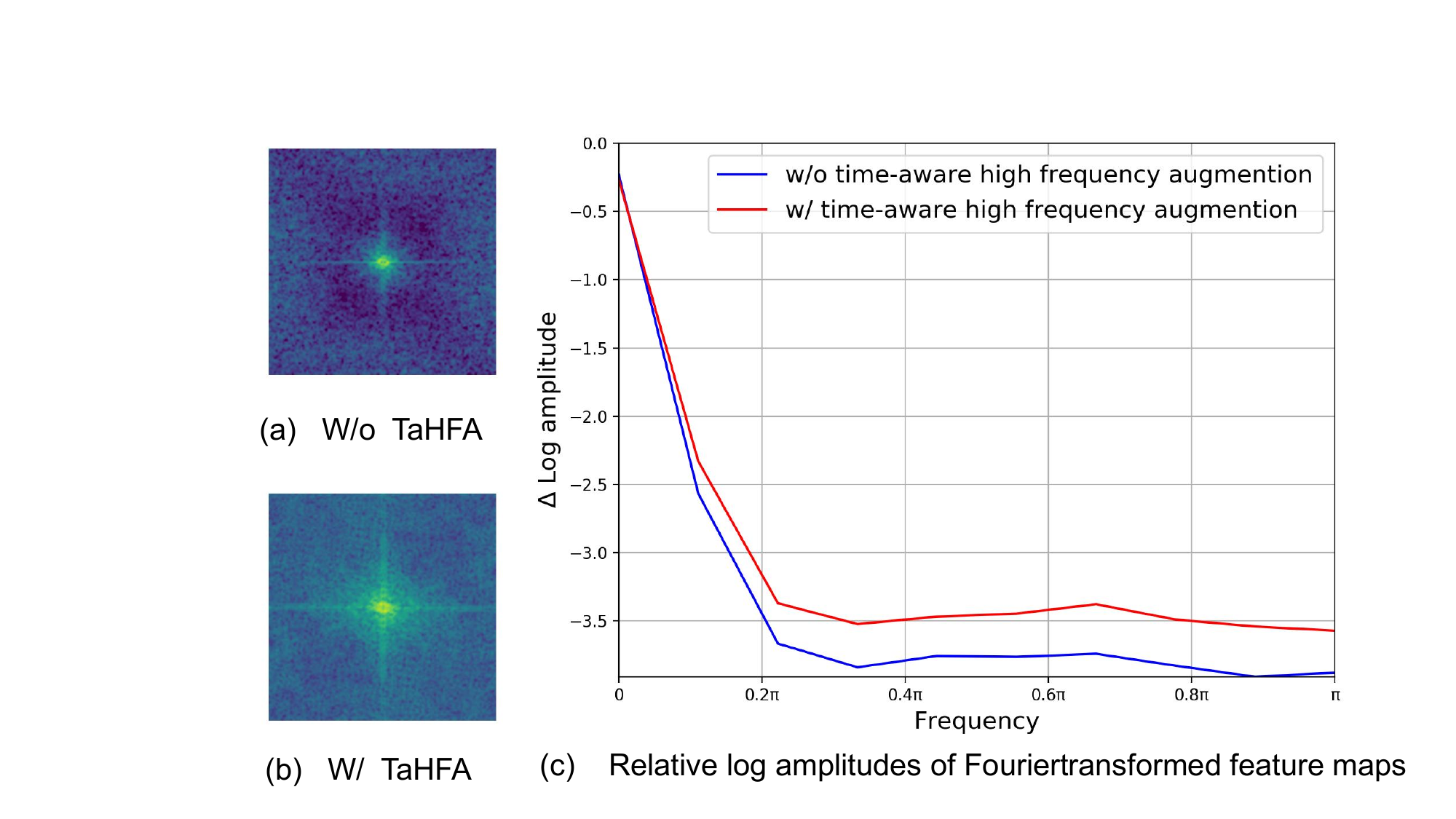}
    \caption{(a)-(b): Fourier spectrum of W and W/o TaHFA. (c): Relative log amplitudes of Fourier-transformed feature maps $\boldsymbol{z}_{\boldsymbol{y}}$.} 
    \label{pic:fig3_tahfa}
\end{figure}
We compute the variance of the $\boldsymbol{y}$ and use the variance to calculate a weighted map $\mathcal{W}(\boldsymbol{y}) = var(\boldsymbol{y})^{-1}$. Simultaneously, to ensure that the high-frequency signals decoded map to image space with features close to the original facial image $\boldsymbol{x}$, we constrain the learning process of the compressor $\mathcal{H}$ through a Face Landmarks Encoder ($\mathbf{E}_{fl}$). With the constraint of landmarks, the parts that can mark facial features can be well emphasized and preserved. Overall, the guiding loss of the compressor is defined as:
\begin{equation}
\resizebox{0.9\linewidth}{!}{$
\begin{aligned}
\mathcal{L}_{\mathcal{H}}=\gamma\mathcal{W}(\boldsymbol{y})\|\boldsymbol{y}-\boldsymbol{z}_{\boldsymbol{y}}\|+\|\mathbf{E}_{fl}(\mathcal{D}(\boldsymbol{z}_{\boldsymbol{y}}))-\mathbf{E}_{fl}(\boldsymbol{x})\|+\lambda\mathcal{R}(\lceil\boldsymbol{y}\rfloor).
\end{aligned}$}
\end{equation}
Here, $\mathcal{R}(\cdot)$ represents the bitrate, and $\lambda,\gamma$ are hyperparameters that adjust the rate-distortion trade-off.
\subsection{Time-Aware High-Frequency Augmentation}
\label{sec:TaHFA}
Solely guiding the compressor through regularization is insufficient; we need the compressor to capture more high-frequency signals in joint training with the control module, ensuring that these signals are preserved in the generation process. To achieve this, we have devised a Time-aware High-Frequency Augmentation (TaHFA), as shown in \cref{pic:pipeline}. The high-frequency control signal $\boldsymbol{z}_{\boldsymbol{y}}$ is fused into the denoising UNet decoder part through the control module. During denoising, the $l$-th decoder layer receives low-frequency interference $\boldsymbol{h}_{l}$ passed through skip connections from the encoder, leading the compressor to learn unnecessary low-frequency components during optimization. To address this issue, we employ spectral modulation in the Fourier domain \cite{si2024freeu} to reduce these low-frequency components (depicted as TaHFA I):
\begin{eqnarray}
\boldsymbol{h}_{l}^{\prime}  =\operatorname{IFFT}\left(\operatorname{FFT}\left(\boldsymbol{h}_{l}\right)\odot \boldsymbol{\beta}_{l}\right).
\end{eqnarray}
Here, $\operatorname{FFT}(\cdot)$ and $\operatorname{IFFT}(\cdot)$ represent the Fourier transform and the inverse Fourier transform, respectively. $\odot$ denotes element-wise multiplication, and $\boldsymbol{\beta}_{l}$ is a mask that actively scales the low-frequency part of $\boldsymbol{h}_{l}$ through a hyperparameter.

Furthermore, as the diffusion model focuses on reconstructing image details in later diffusion steps, we aim to enhance high-frequency image control to adapt to the denoising patterns in the later diffusion steps. Specifically, during training, we blend $\boldsymbol{z}_{\boldsymbol{y}}$ and $\boldsymbol{y}$ at different time steps in varying proportions (depicted as TaHFA II):
\begin{eqnarray}
\boldsymbol{z}^{\prime} _{\boldsymbol{y}}(t)=\left(\sqrt{1-t/T}\right)\boldsymbol{z}_{\boldsymbol{y}}+\left(1-\sqrt{1-t/T}\right)\boldsymbol{y}.
\end{eqnarray}
$\boldsymbol{z}^{\prime} _{\boldsymbol{y}}(t)$ avoids the compressor being influenced by early time steps and instead focuses on the denoising process in later time steps. As shown in the \cref{pic:fig3_tahfa}, TaHFA enables the compressor to capture more high-frequency signals. Further experiments indicate that these high-frequency signals enhance the realism of decoded images and strengthen support for vision tasks.

\subsection{Hybrid Low-Frequency Enhancement}
\label{sec:HLE}
Relying solely on high-frequency signal control can diminish the realism and stability of generated images \cite{careil2023towards}, as confirmed by our experiments. However, CLIP-based weak alignment training extracts vague semantic information that struggles with pixel-level restoration tasks \cite{wang2024instantid}. To address this issue, we introduce the Hybrid Low-Frequency Semantic Control Module, as shown in the HLFE module in \cref{pic:pipeline}.

In this phase, we utilize a pre-trained face embedding encoder $\mathbf{E}_{fe}$ to extract facial semantic embeddings from $\hat{\boldsymbol{x}}$ and generate the mixed low-frequency control signal $\boldsymbol{s}$ jointly with decoupled cross-attention and text semantic embeddings through a mapping layer.

The control module based on the control network is sensitive to high-frequency control but overlooks high-level information such as color and style. This results in the underutilization of low-frequency components in $\boldsymbol{z}_{\boldsymbol{y}}$. Essentially, we decouple the low-frequency components of the compressor's decoding information and integrate a more semantically rich and nuanced prompt hint set under the guidance of text embeddings.

\subsection{Training Strategies}
Our training is divided into two stages. In the first stage, we aim to optimize the compressor $\mathcal{H}$ and the control module. During this phase, we do not introduce low-frequency control signals to enable $\mathcal{H}$ to achieve a better balance between high-frequency capturing capability and bit rate in joint training. The optimization objective in this stage is:
\begin{equation}
\resizebox{0.8\linewidth}{!}{$
\begin{aligned}
&\mathcal{L}_{stage\ 1}=\mathcal{L}_{\mathcal{H}}+\mathcal{L}_{LDM},
\\
&\mathcal{L}_{LDM}=\mathbb{E}_{\boldsymbol{z}_t,t, \epsilon}\left[\left\|\epsilon-\mathcal{M}_{\theta}\left(\boldsymbol{z}_t,\boldsymbol{z}^{\prime}_{\boldsymbol{y}}(t), t\right)\right\|_2^2\right].
\end{aligned}$}
\end{equation}

\begin{figure}[t!]
    \centering
        \begin{minipage}[t]{\columnwidth}
        \begin{algorithm}[H]
            \caption{Training Stage I}
            \begin{algorithmic}[1]
                \STATE Given input data $\boldsymbol{x}$, Stable Diffusion en/decoder $\mathcal{E,D}$, compressor $\mathcal{H}_{\phi}$, control module $\mathrm{CM}_{\gamma}$, learning rate $\varepsilon$.
                \REPEAT
                \STATE $t \sim \mathcal{U}\left(0,1,2, . ., T)\right)$
                \STATE $\epsilon \sim \mathcal{N}(\mathbf{0}, \mathbf{I})$
                \STATE $\boldsymbol{y},e_1,e_2=\mathcal{E}(\boldsymbol{x})$
                \STATE $\mathcal{W}(\boldsymbol{y}) = var(\boldsymbol{y})^{-1}$
                \STATE $\boldsymbol{z}_t=\sqrt{\overline{\alpha}_t} \boldsymbol{y}+\sqrt{1-\overline{\alpha}_t} \epsilon$
                \STATE $\boldsymbol{z}_{\boldsymbol{y}},\lceil \boldsymbol{y} \rfloor=\mathcal{H}_{\phi}(\boldsymbol{y},e_1,e_2)$
                \STATE $\mathcal{L}_{\mathcal{H}_{\phi}}=\gamma\mathcal{W}(\boldsymbol{y})\|\boldsymbol{y}-\boldsymbol{z}_{\boldsymbol{y}}\|+\|\mathbf{E}_{fl}(\mathcal{D}(\boldsymbol{z}_{\boldsymbol{y}}))-\mathbf{E}_{fl}(\boldsymbol{x})\|+\lambda\mathcal{R}(\lceil\boldsymbol{y}\rfloor)$
                \STATE $\boldsymbol{z}^{\prime} _{\boldsymbol{y}}(t)=\left(\sqrt{1-t/T}\right)\boldsymbol{z}_{\boldsymbol{y}}+\left(1-\sqrt{1-t/T}\right)\boldsymbol{y}$
                \STATE $c_{image}=\mathrm{CM}_{\gamma}(\boldsymbol{z}^{\prime} _{\boldsymbol{y}}(t),t)$
                \STATE $\mathcal{L}_{ldm}=\left\|\epsilon-\mathcal{M}_{\theta}\left(\boldsymbol{z}_t,c_{image}, t\right)\right\|_2^2$
                \STATE $\mathcal{L}_{1st}=\mathcal{L}_{ldm}+\mathcal{L}_{\mathcal{H}}$
                \STATE $(\theta, \gamma,\phi)=(\theta,\gamma, \phi)-\varepsilon \nabla_{\theta,\gamma, \phi} \mathcal{L}_{1st}$
                \UNTIL{converge}
            \end{algorithmic}
        \end{algorithm}
         \end{minipage}
        \hspace{.1in}
      \begin{minipage}[t]{\columnwidth}
        \centering
         \begin{algorithm}[H]
                \caption{Training Stage II}
                \begin{algorithmic}[1]
                    \STATE Given input data $\boldsymbol{x}$, Stable Diffusion en/decoder $\mathcal{E,D}$, compressor $\mathcal{H}_{\phi}$, control module $\mathrm{CM}_{\gamma}$, learning rate $\varepsilon$.
                    \REPEAT
                    \STATE $text=\mathrm{IC}(\boldsymbol{x})$
                    \STATE $s_t=\mathbf{E}_{te}(text)$
                    \STATE $t \sim \mathcal{U}\left(0,1,2, . ., T)\right)$
                    \STATE $\epsilon \sim \mathcal{N}(\mathbf{0}, \mathbf{I})$
                    \STATE $\boldsymbol{y},e_1,e_2=\mathcal{E}(\boldsymbol{x})$
                    \STATE $\boldsymbol{z}_t=\sqrt{\overline{\alpha}_t} \boldsymbol{y}+\sqrt{1-\overline{\alpha}_t} \epsilon$
                    \STATE $\boldsymbol{z}_{\boldsymbol{y}}=\mathcal{H}_{\phi}(\boldsymbol{y},e_1,e_2)$
                    \STATE $\hat{\boldsymbol{x}}=\mathcal{D}(\boldsymbol{z}_{\boldsymbol{y}})$
                    \STATE $s_f=\mathrm{Proj}(\mathbf{E}_{fe}(\hat{\boldsymbol{x}}))$
                    \STATE $s=\mathbf{DCS}(s_t,s_f)$
                    \STATE $c_{image}=\mathrm{CM}_{\gamma}(\boldsymbol{z}_{\boldsymbol{y}},t)$
                    \STATE $\hat{\boldsymbol{y}}=\mathrm{Sampler}(\mathcal{M}_{\theta}\left(c_{image}, s\right),\epsilon,steps=3)$
                \STATE $\mathcal{L}^{\prime}_{ldm}=\left\|\epsilon-\mathcal{M}_{\theta}\left(\boldsymbol{z}_t,c_{image}, s,t\right)\right\|_2^2$
                    \STATE $\mathcal{L}_{2st}=\|\hat{\boldsymbol{y}}-\boldsymbol{y}\|_2+\operatorname{LPIPS}(\mathcal{D}(\hat{\boldsymbol{y}})-\boldsymbol{x})+\mathcal{L}^{\prime}_{ldm}$
                    \STATE $(\theta, \gamma)=(\theta,\gamma)-\varepsilon \nabla_{\theta,\gamma} \mathcal{L}_{2st}$
                    \UNTIL{converge}
                \end{algorithmic}
            \end{algorithm}
        \end{minipage}
\label{alg:appendix_alg_2}
\end{figure}
During the training of the second stage, we aim to stabilize the denoising model's generation of high-frequency details and global color information under semantic guidance. To achieve this, we freeze the parameters in $\mathcal{H}$ to obtain a stable $\hat{\boldsymbol{x}}$. Simultaneously, we unfreeze the cross-attention layer in the denoising U-Net to adapt to new semantic embeddings during training. Lastly, for better control over the training outcomes, we constrain the preliminary sampled results $\hat{\boldsymbol{y}}$ in both the latent space and pixel space to align more closely with the input image:
\begin{equation}
\resizebox{0.85\linewidth}{!}{$
\begin{aligned}
    &\mathcal{L}_{stage\ 2}=\|\hat{\boldsymbol{y}}-\boldsymbol{y}\|_2+\operatorname{LPIPS}(\mathcal{D}(\hat{\boldsymbol{y}})-\boldsymbol{x})+\mathcal{L}^{\prime}_{LDM},\\
&\mathcal{L}^{\prime}_{LDM}=\mathbb{E}_{\boldsymbol{z}_t,\boldsymbol{z}_{\boldsymbol{y}},\boldsymbol{s},t, \epsilon}\left[\left\|\epsilon-\mathcal{M}_{\theta}\left(\boldsymbol{z}_t,\boldsymbol{z}_{\boldsymbol{y}},\boldsymbol{s}, t\right)\right\|_2^2\right].
\end{aligned}$}
\end{equation}
where $\operatorname{LPIPS}$ denotes the LPIPS loss \cite{zhang2018unreasonable}. Constraining $\hat{\boldsymbol{y}}$ requires only rough sampled results. Therefore, we set the sampling steps to obtain $\hat{\boldsymbol{y}}$ fixed at 3 during training.

\subsection{Detailed Algorithm}
\begin{figure}[t!]
    \centering
        \begin{minipage}[t]{\linewidth}
        \begin{algorithm}[H]
            \caption{Encode Stage}
            \begin{algorithmic}[1]
                \STATE Given input data $\boldsymbol{x}$, Stable Diffusion encoder $\mathcal{E}$, compressor $\mathcal{H}_{\phi}$.
                \STATE $text=\mathrm{IC}(\boldsymbol{x})$
                \STATE $\boldsymbol{y},e_1,e_2=\mathcal{E}(\boldsymbol{x})$
                \STATE $\lceil \boldsymbol{y} \rfloor=\mathcal{H}^{e}_{\phi}(\boldsymbol{y},e_1,e_2)$
                \STATE Encode $\lceil \boldsymbol{y} \rfloor$, $text$ to binary file
                \STATE Output encoded data
            \end{algorithmic}
        \end{algorithm}
         \end{minipage}
  \begin{minipage}[t]{\linewidth}
    \centering
 \begin{algorithm}[H]
            \caption{Decode Stage}
            \begin{algorithmic}[1]
                \STATE Given encoded data  $\lceil \boldsymbol{y} \rfloor$, $text$, Stable Diffusion decoder $\mathcal{D}$, control module $\mathrm{CM}_{\gamma}$.
                \STATE $s_t=\mathbf{E}_{te}(text)$
                \STATE $\boldsymbol{z}_{\boldsymbol{y}}=\mathcal{H}^d_{\phi}(\lceil \boldsymbol{y} \rfloor)$
                \STATE $\boldsymbol{z}_T=\sqrt{\overline{\alpha}_T} \boldsymbol{z}_{\boldsymbol{y}}+\sqrt{1-\overline{\alpha}_T} \epsilon$
                \STATE $\hat{\boldsymbol{x}}=\mathcal{D}(\boldsymbol{z}_{\boldsymbol{y}})$
                \STATE $s_f=\mathrm{Proj}(\mathbf{E}_{fe}(\hat{\boldsymbol{x}}))$
                \STATE $s=\mathbf{DCS}(s_t,s_f)$
                \STATE $c_{image}=\mathrm{CM}_{\gamma}(\boldsymbol{z}_{\boldsymbol{y}},t)$
                \STATE $\hat{\boldsymbol{y}}=\mathrm{Sampler}(\mathcal{M}_{\theta}\left(c_{image}, s\right),\boldsymbol{z}_T,steps)$
                \STATE Output $\boldsymbol{x}_{rec}=\mathcal{D}(\hat{\boldsymbol{y}})$
            \end{algorithmic}
        \end{algorithm}
          \end{minipage}
\label{alg:appendix_alg_1}
\end{figure}
We provide pseudocode for the first and second stages of training, as well as a demonstration of the encoding and decoding processes during inference, as shown in Alg.1 to Alg.4. Here, $\mathbf{E}_{fl}$ and $\mathbf{E}_{fe}$ represent the pre-trained landmark encoder and facial feature encoder, respectively. $\mathcal{H}^{e}_{\phi}$ and $\mathcal{H}^{d}_{\phi}$ denote the encoder and decoder of the compressor $\mathcal{H}_{\phi}$. $\mathrm{IC}$ represents the image caption model. $\mathrm{Proj}$ signifies the non-linear modulation layer, and $\mathrm{DCS}$ stands for decoupled cross-attention. Additionally, we further define a sampler as $\mathrm{Sampler}(\cdot,\cdot,\cdot)$, which takes the denoising network $\mathcal{M}_{\theta}$, initial input $\boldsymbol{z}_t$, and the number of sampling steps as inputs.
\section{Experiments}
\label{sec:experiments}

\begin{figure*}[t]
    \centering
    \includegraphics[width=1\linewidth]{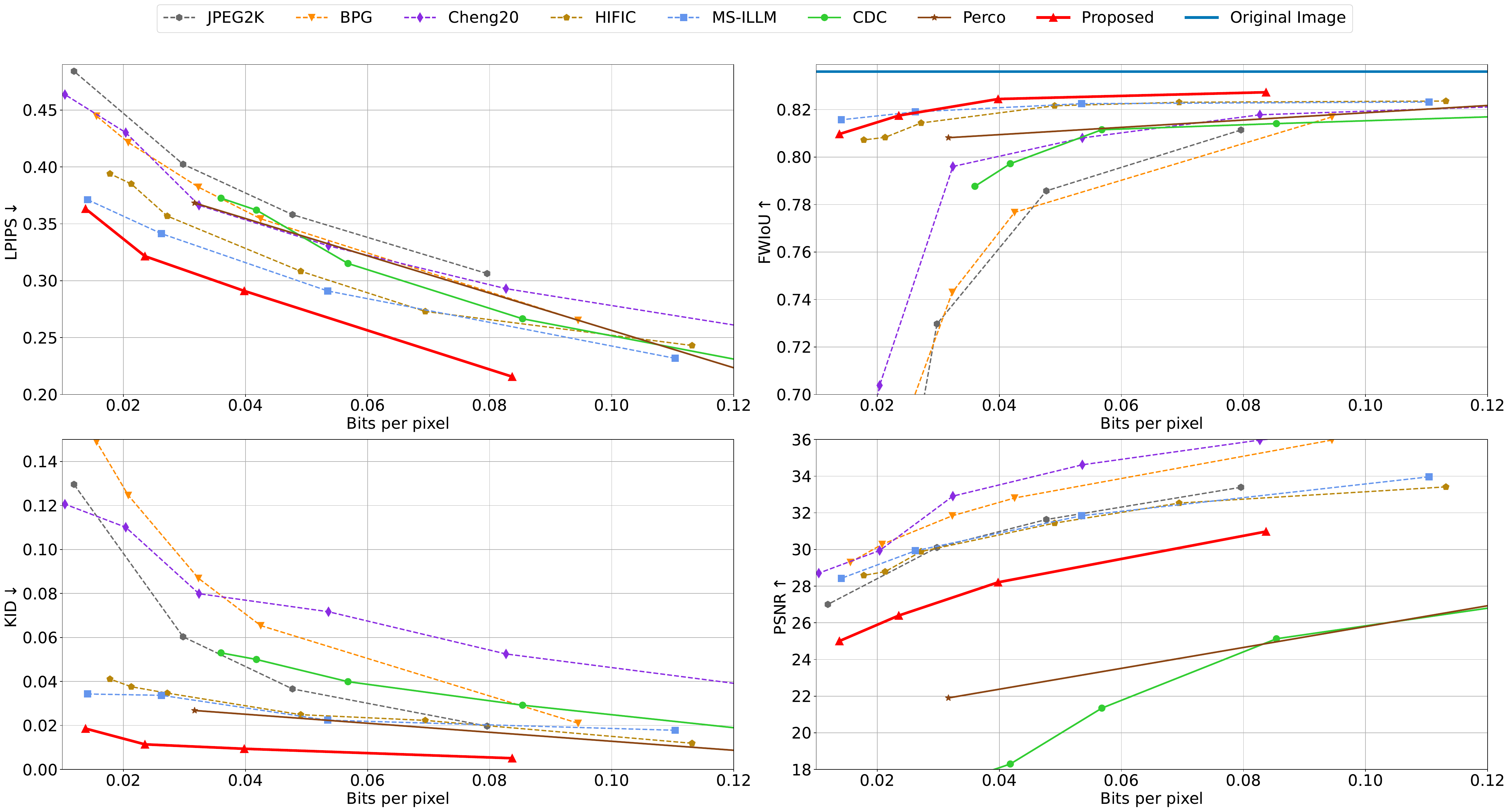}
    \caption{RD-performance between bitrates and metrics, including human vision metric LPIPS, KID and PSNR, and machine vision metric FWIoU. The $\uparrow$ or $\downarrow$ means higher or lower is better respectively.} 
    \label{pic:rd-regular}
\end{figure*}
% \begin{figure*}
%     \centering
%     \includegraphics[width=1.0\linewidth]{assets/plot_face_m_try.pdf}
%     \caption{RD-performance between bitrates and different downstream face-related tasks, include the facial image segmentation and gender classification. The $\uparrow$ or $\downarrow$ means higher or lower is better respectively.} 
%     \label{pic:rd-rask}
% \end{figure*}
\subsection{Experimental Setting}
\subsubsection{Training Details}
We employed the pre-trained Stable Diffusion 2.1 \footnote{https://huggingface.co/stabilityai/stable-diffusion-2-1-base}
% \footnote{Stable Diffusion v2.1\url{}} 
as the base model, which is frozen during training to preserve its generative ability and reduce training costs. We trained the proposed method on the FFHQ dataset\cite{ffhq}, which consists of 70,000 facial images in $1024 \times 1024$. We randomly crop each image to 256 × 256 resolution for efficient training. We implemented our model in the PyTorch\cite{pytorch} framework and trained it on a single Nvidia A6000 GPU. We used the Adam\cite{kingma2014adam} optimizer with a learning rate of $1e^{-4}$ at the first stage,and $0.5e^{-5}$ at the second stage. We maintained $\gamma=0.2$ at a fixed value and modulated the bitrate size by adjusting $\lambda$. Our $\lambda$ values were set as $\lambda=\{32, 96, 190, 224\}$. Besides, we have opted for \cite{wang2019adaptive} as the pre-trained facial landmark extractor. Drawing inspiration from \cite{si2024freeu}, we slightly elevated the recommended values by setting $\beta_l$ to $0.6$. For the training of diffusion in \cref{sec:HLE}, the total step length $T$ was configured to be $1000$.
Furthermore, we used the model \cite{deng2018arcface} in the InsightFace project \footnote{InsigntFace:\url{https://github.com/deepinsight/insightface}} as the pre-trained facial feature extractor, and BLIP2\cite{li2023blip} as our image captioning model.

\subsubsection{Evaluation}
\noindent \textbf{Datasets.}
We tested our method on the CelebA-HQ test dataset\cite{CelebAMask-HQ}, the same as in previous work. Moreover, to verify the generalization, we also tested methods on the Facescrub dataset\cite{Ng2014ADA}, which had not been tested in previous work. The CelebA-HQ test dataset consists of 2,824 facial images with a resolution of $1024\times 1024$, while the Facescrub dataset consists of 436 high-resolution facial images. During the testing process, we resized these images to a resolution of  $1024\times 1024$. For FID and KID evaluations, in order to conduct a more stable test, we used a subset of CelebA, which consists of 1,000 facial images in $256\times 256$ in the PNG format.
\noindent \textbf{Metrics.} 
Multiple evaluation metrics were employed to fully assess the performance of the model. Similar to other compression tasks, we used bits per pixel (bpp) as a metric to measure the degree of compression. Based on the evaluation types, metrics can be categorized into five classes.
\textit{(1) distortion-based metrics}: PSNR. This metric compares the differences between each image pixel by pixel, which is hard to reflect the reconstruction quality of face images in human vision and downstream tasks.
\textit{(2) Reference-based perceptual-based metrics}: LPIPS\cite{zhang2018unreasonable} and DISTS\cite{ding2020image}. These metrics can effectively reflect the overall image quality and the reconstruction performance as perceived by human vision.
\textit{(3) Generative model perceptual similarity metrics}: FID\cite{heusel2017gans} and KID\cite{binkowski2018demystifying}. These metrics place greater emphasis on evaluating the visual effects, content, and structural aspects of the images.
\textit{(4) self-evaluation perceptual-based metrics}: CLIP-IQA \cite{wang2022exploring} and FS \cite{liao2024facescore}. These metrics leverage existing pre-trained models to evaluate whether the generated images maintain semantic consistency. Specifically, FS employs an image inpainting pipeline based on a diffusion model that has been fine-tuned with ImageReward \cite{xu2024imagereward}, thereby measuring the facial quality of the generated images.
\textit{(5) downstream-task-based metrics}: FWIoU \cite{garcia2017review} and gender-accuracy. These metrics are used to measure the accuracy of the reconstructed images in downstream tasks, and they respectively correspond to face segmentation and gender classification.

To ensure consistency and reliability, we adopted the official libraries \footnote{FWIoU: \url{https://github.com/switchablenorms/CelebAMask-HQ}}\footnote{Gender Classification: \url{https://github.com/ndb796/CelebA-HQ-Face-Identity-and-Attributes-Recognition-PyTorch}}\footnote{FS: \url{https://github.com/OPPO-Mente-Lab/FaceScore}} for evaluation metrics. Additionally, for each evaluation method, we utilized the checkpoints supplied by the official repositories to perform the assessments. For LPIPS, we utilized the \textit{lpips} library. For DISTS, we used the \textit{dists}. As for FID and KID, \textit{clean-fid} library was used with default settings. For CLIP-IQA, we used the \textit{pyiqa} library.

\subsubsection{Baselines}
To demonstrate and validate the effectiveness of our proposed method, we compared it against the current state-of-the-art and most widely adopted deep image compression techniques. Specifically, the baseline methods included the widely used classical compression methods JPEG2000 \cite{jpeg2000} and BPG, DNN-based model Cheng20 \cite{cheng2020learned}, GAN-based models HiFiC \cite{hific} and MS-ILLM \cite{illm}, as well as the diffusion-based model CDC \cite{yang2024lossy} and Perco \cite{careil2023towards}. We meticulously adjusted the hyper-parameters to ensure that the compression performance of each baseline model operates within the bpp range of 0.01 to 0.1. 
To ensure fairness, all baselines except for the Perco method are retrained on the same training set as our proposed method, with hyperparameters aligned as closely as possible to those in the original papers. Due to computational constraints, we were unable to retrain the Perco method; however, its provided checkpoints are trained on the Open Images V6 dataset, which includes 1,743,042 images. We believe that comparing results trained on such a large dataset is reasonably fair.

% This rigorous comparison allows us to objectively assess the relative performance and advantages of our method in the context of contemporary facial image compression frameworks.

\subsection{Rate-Distortion Performance}
% Fig.\ref{pic:rd-regular} shows the comparison results of our proposed method with the baselines in the reference-based perceptual-based metric LPIPS and the no-reference perceptual-based metric FID. 
\cref{pic:rd-regular} shows the comparison results of our proposed method with the baselines in the human vision metrics LPIPS, KID and PSNR and the machine vision metric FWIoU.

As depicted in the \cref{pic:rd-regular}, our proposed method surpasses various existing approaches across perceptual metrics. This improvement is attributable to our model's ability to retain low-frequency information at low bitrates while specifically optimizing the storage of essential high-frequency details related to facial features. Consequently, the reconstructed images not only maintain semantic consistency but also achieve perceptual fidelity that more closely aligns with the original images.
% On the other hand, for the indicators of distortion-based, the indicators of the proposed method are relatively low.This is because distortion-based metrics, like MS-SSIM, is based on the comparison of similarity between pixels. At low bit rates, traditional methods are more inclined to retain the original values of pixels or discard the values that can be predicted from surrounding pixels. This makes the distortion-based metrics high, but the actually presented image is blurry. However, the proposed method uses the diffusion model to refine these blurs, making the reconstructed image more realistic and thus performing better in the perceptual-based metrics, like LPIPS or FID. 
For a more detailed and qualitative analysis of the compression results, please refer to the section \ref{sec:visual_results}

Furthermore, to demonstrate the application of the proposed method in downstream face-related tasks, we select the face segmentation task. In face-related downstream tasks, our proposed method outperforms CNN-based approaches at low bit rates. And the images obtained by our method demonstrate performance in downstream models that is highly comparable to that of the original images. Thus, it can be inferred that our method can essentially preserve the performance of machine vision under low bit rates. Additionally, in fine-grained tasks, such as facial segmentation, the model proposed by us enhances high-frequency information. Consequently, it attains an optimal performance that is closer to that of the original images. 

In \cref{table:q}, we selected several points with comparable bpp at low bit rates and presented detailed metrics across various tasks. The results demonstrate that our method performs exceptionally well in both human visual perception and machine vision tasks. For certain methods, we selected two models with similar bpp for comparison to ensure a more reasonable and fair evaluation. As a result, some methods like Cheng20, HiFiC, CDC and Ours are presented with two rows of data in \cref{table:q}.

Furthermore, to demonstrate the generalization ability of the proposed method, we conducted further experiments on the Facescrub dataset, which had not been tested in previous studies. The results are presented in Table \ref{tab:dataset}. The additional experiments indicate that our method can achieve consistent and outstanding performance on other datasets as well, thus verifying the generalization ability of the proposed approach.

\begin{figure*}[t]
    \centering
    \includegraphics[width=0.98\linewidth]{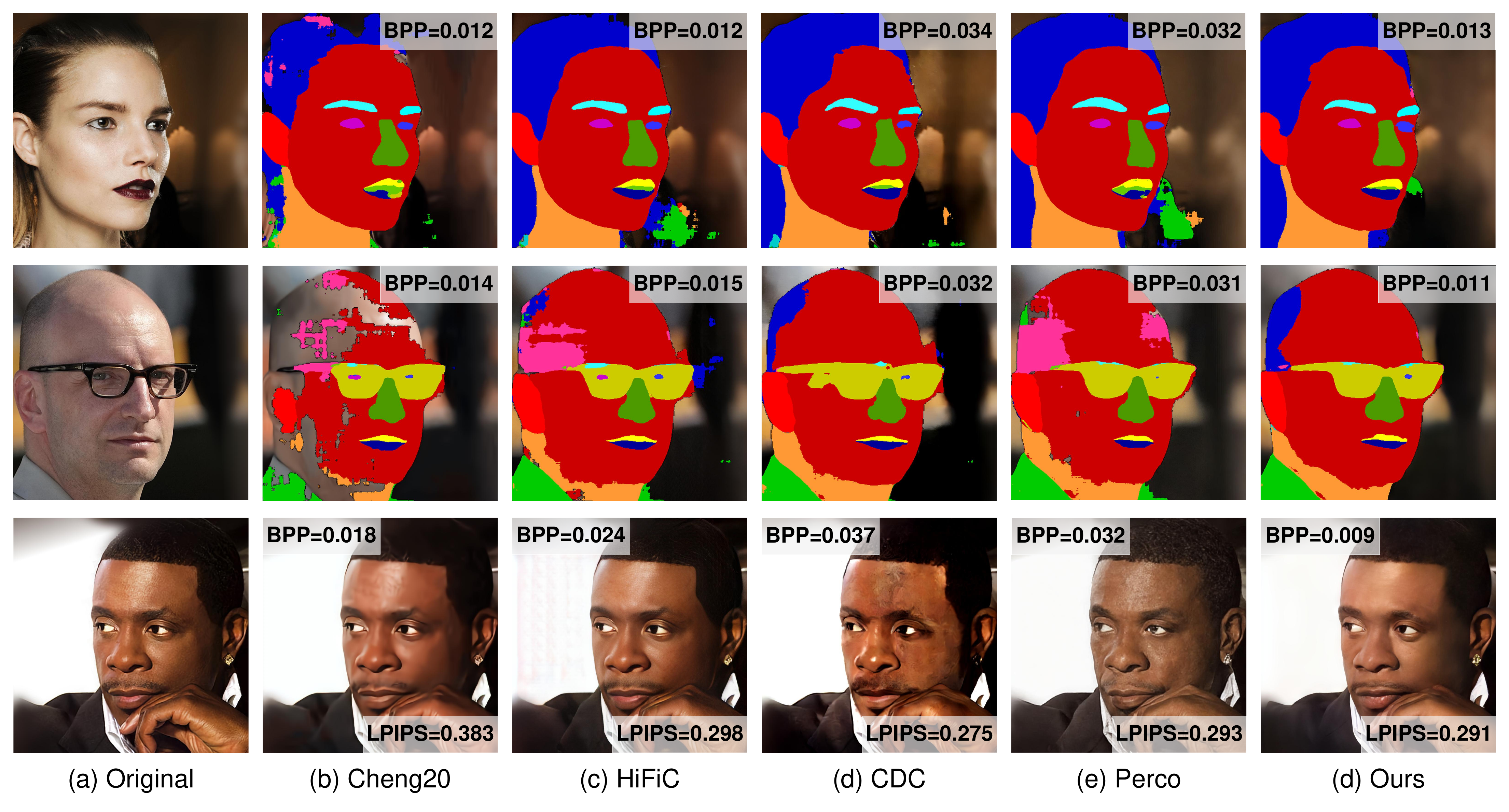}
    \caption{The original Images and the decompressed images of different baselines on the CelebA-HQ dataset. For each image, the upper corner is labeled as bpp. For human-vision visualization, the  bottom-right corner is labeled as LPIPS. Our proposed method demonstrates superior performance in image construction at significantly lower bpp.} 
    \label{pic:vis-pic}
\end{figure*}
% on one hand, the performance bottlenecks of downstream task models limit the potential enhancements that compression algorithms can contribute to these tasks. On the other hand, most existing face downstream task algorithms initially crop face images to smaller sizes to facilitate face alignment, which prevents the effective transmission of additional fine-grained details to the downstream models. 

% \begin{figure}[t]
%     \centering
%     \includegraphics[width=1.0\linewidth]{assets/visualcompare_H.pdf}
%     \caption{Decompressed Images from CelebA-HQ based on baselines. For each image, the upper-left corner is labeled as bpp and the lower-right corner is labeled as LPIPS.} 
%     \label{pic:vis-regular}
% \end{figure}
\begin{table}
\caption{Quantitative Evaluation Results on the CelebA-HQ dataset at comparable bpp. Blod highlights the best outcomes, For certain models, we conducted evaluations at multiple compression rates to achieve a more comprehensive comparison.}
\centering
\resizebox{1\linewidth}{!}{
\begin{tblr}{
  cells = {c},
  cell{1}{1} = {r=2}{},
  cell{1}{2} = {r=2}{},
  cell{1}{4} = {c=2}{},
  cell{1}{6} = {c=2}{},
  cell{3}{1} = {r=2}{},
  cell{5}{1} = {r=2}{},
  cell{7}{1} = {r=3}{},
  cell{5}{2} = {r=2}{},
  cell{7}{2} = {r=2}{},
  cell{10}{2} = {r=2}{},
  cell{13}{2} = {r=2}{},
  cell{10}{1} = {r=5}{},
  vline{2,3,4,6} = {1,2,4,6,8,9,10,11,12,13,14,16}{},
  vline{4,6} = {2}{},
  vline{2-4,6} = {2,3,5,7,10}{},
  vline{3-4,6} = {2,3,4,6,8-9,11-14}{},
  hline{1,15} = {-}{0.08em},
  hline{2} = {3-7}{},
  hline{3,5,7,10} = {-}{},
  hline{4,9,12,13} = {2-7}{},
}
Category        & Method     & Bit rate                & Human Vision            &                        & Machine Vision          &                          \\
                &            & BPP $\downarrow$        & CLIP-IQA $\uparrow$     & FS $\uparrow$          & FWIoU $\uparrow$        & Gender $\uparrow$        \\
Traditional     & JPEG 2000  & 0.029~                  & 0.228~                  & 3.44~                  & 0.729~                  & 97.83\%                  \\
                & BPG        & 0.021~                  & 0.186~                  & 3.21~                  & 0.662~                  & 97.69\%                  \\
DNN based       & Cheng20 & 0.020~                  & 0.329~                  & 3.40~                  & 0.704~                  & 96.92\%                  \\
                & Cheng20& 0.032~                  & 0.384~                  & 3.67~                  & 0.574~                  & 94.90\%                  \\
GAN based       & HiFiC      & 0.018~ & 0.559~                  & 4.55~                  & 0.808~                  & 99.10\%                  \\
                & HiFiC      & 0.021~                  & 0.545~                  & 4.56~                  & 0.807~                  & 99.22\%                  \\
                & MS-ILLM    & 0.026~                  & 0.531~                  & 4.57~                  & 0.816~ & \textbf{99.58\%}         \\
Diffusion based & CDC        & 0.036~                  & 0.450~                  & 1.62~                  & 0.788~                  & 82.83\%                  \\
                & CDC        & 0.042~                  & 0.459~                  & 2.46~                  & 0.797~                  & 91.18\%                  \\
                & Perco      & 0.032~                  & 0.484~                  & 4.05~                  & 0.808~                  & 98.12\%                  \\
                & Ours       & \textbf{0.013~}         & \textbf{0.580~}         & 4.65~ & 0.810~                  & 98.60\%                  \\
                & Ours       & 0.024~                  & 0.577~ & \textbf{4.67~}         & \textbf{0.817~}         & 99.29\% 
\end{tblr}}
\label{table:q}
\end{table}

\subsection{Visual Results}
\label{sec:visual_results}
% fig.\ref{pic:vis-regular} shows the comparison of examples of reconstruction results of our proposed method and the baseline. For each image, the upper left corner of the image indicates the bpp of the single-image phenomenon, and the lower right corner of the image indicates the LPIPS of the single image.

% The results demonstrate that DNN-based methods, such as Cheng20, effectively preserve the overall structure of the original image but fail to accurately reconstruct high-frequency details, such as strands of hair. In contrast, GAN-based approaches like HiFiC enhance high-frequency information through generative modeling; however, they produce noticeable false contour in homogeneous regions, such as solid color backgrounds or facial areas, due to inherent limitations of GAN architectures. Additionally, existing diffusion-based compression methods, taking CDC and Perco for examples, which are not specifically designed for face images, exhibit color discrepancies during image reconstruction. In comparison, our proposed method maintains low-frequency features and refines high-frequency details at low bitrates relative to the baseline, resulting in images that are more realistic and closely resemble the original.

% Similarly, 
We visualize the segmentation results and human-vision results as shown in \cref{pic:vis-pic}. 

For segmentation results, due to the inferior reconstruction performance of Cheng20 and CDC in low bitrates, downstream segmentation models struggle to accurately identify various parts of the images, resulting in unrecognizable regions. HiFiC, due to the erroneous enhancement of high-frequency details, introduces artifacts that lead to incorrect segmentation in areas outside the facial regions. Although Perco achieves relatively good segmentation outcomes, it still experiences segmentation errors in certain areas influenced by the background environment for overly relying on low-frequency information. Thanks to its ability to capture and enhance high-frequency information, the proposed method achieves more accurate segmentation results compared to the baseline, and avoids erroneous segmentation in non-facial regions.

In terms of human visual perception, our method successfully addresses the artifact introduction issue prevalent in previous approaches. For instance, the HiFiC method exhibits noticeable noise and regular artifacts in extensive blank background regions. While maintaining certain fidelity in facial areas, this approach results in visually inconsistent reconstruction across the entire image, significantly compromising the overall perceptual quality. Furthermore, by effectively incorporating and enhancing frequency domain information, the proposed approach achieves remarkable reconstruction quality even at ultra lower bit rates.

\begin{table}[]
\centering
\caption{Extention Experiments on FaceScrub Dataset. Bold highlights the best outcomes.$\downarrow$ or $\uparrow$ represent lower or higher is better respectively}
\begin{tabular}{ccccc}
\toprule
Method  & BPP $\downarrow$ & LPIPS $\downarrow$ & FID $\downarrow$  &FWIoU $\uparrow$\\ 
\midrule
Cheng20 &  0.048   &0.415  &88.97   &0.714 \\
HiFiC & 0.043  &0.346  &30.44   & 0.836 \\
MS-ILLM & 0.051  & 0.328 &18.75   &\textbf{0.858}\\
CDC& 0.039   & 0.396 &78.40   &0.763 \\
Ours    & \textbf{0.038}    &\textbf{0.298}  &\textbf{16.02}   &\textbf{0.858} \\
\bottomrule
\end{tabular}
\label{tab:dataset}
\end{table}
% \begin{figure*}[t]
%     \centering
%     \includegraphics[width=0.91\linewidth]{assets/visualcompare_M2.pdf}
%     \caption{The original Images and the segmentation results of decompressed images of different baselines on the CelebA-HQ dataset. For each image, the upper-right corner is labeled as bpp.} 
%     \label{pic:vis-seg}
% \end{figure*}

\subsection{Ablation Study}
We conducted ablative experiments on the various modules proposed, as shown in the \cref{tab:ablation}. We utilized BD-rate \cite{bjontegaard2008improvements} as a metric to gauge the extent of decrease (or increase) in bit rate at the same level of distortion compared to the reference point.  \textbf{0\%} represents the reference point, and other positive values represent the performance degradation caused by the absence of corresponding modules. Setting the complete framework as the reference point, it is evident that all ablation models exhibited a significant decline in performance. Among them, FCG and HLFE significantly enhance both the machine vision performance and human vision of the decoded images. Although TaHFA shows a relatively modest improvement in comparison, TaHFA does not require the introduction of any additional trainable parameters and does not add to the computational burden during inference.

\begin{table}
\centering

\caption{BD-rate for different methods on the CelebA-HQ. The positive value indicate the ratio of additional bits required to achieve the same LPIPS when specific modules are omitted, relative to the complete method.}
\resizebox{1\linewidth}{!}{
\begin{tblr}{
  cells = {c},
  cell{1}{1} = {r=2}{},
  cell{1}{2} = {c=2}{},
  cell{1}{4} = {c=2}{},
  vline{2} = {1,2}{0.05em},
  vline{4} = {1,2}{0.05em},
  vline{2,4} = {3-7}{0.05em},
  hline{1,8} = {-}{0.08em},
  hline{2} = {2-5}{0.03em},
  hline{3,7} = {-}{0.05em},
}
Model         & Human Vision &            & Machine Vision &            \\
              & LPIPS        & DISTS      & FWIoU          & Gender     \\
W/o FCG ~      & 24.47\%        & 26.32\%      & 121.69\%         & 57.32\%      \\
W/o TaHFA I ~  & 12.89\%        & 13.3\%       & 34.22\%          & 8.96\%       \\
W/o TaHFA II ~ & 11.07\%        & 6.99\%       & 17.29\%          & 15.34\%      \\
W/o HLFE      & 28.93\%        & 40.15\%      & 191.6\%          & 61.44\%      \\
\textbf{Ours} & \textbf{0\%}   & \textbf{0\%} & \textbf{0\%}     & \textbf{0\%} 
\end{tblr}}
\vspace{-10pt}
\label{tab:ablation}
\end{table}
\section{Conclusion}
\label{sec:conclusion}

In this work, we explore the application of diffusion models in the task of face image compression to help achieve better results for compression at lower bit rates. We analyze facial image compression methods from a frequency domain perspective and propose FaSDiff. Specifically, FaSDiff preserves high-frequency signals to enhance the model's performance in machine vision and aligns and enhances low-frequency information to improve perceptual quality for human vision simultaneously. Extensive experiments on human vision and machine vision indicate that FaSDiff shows outstanding performance in both image representation and downstream tasks.

% \section*{Acknowledgments}
% This should be a simple paragraph before the References to thank those individuals and institutions who have supported your work on this article.
\normalem
\bibliographystyle{IEEEtran}
\bibliography{main}

@String(CVPR= {IEEE Conf. Comput. Vis. Pattern Recog.})

@String(ICIP = {IEEE Int. Conf. Image Process.})

@String(AAAI = {AAAI})

@String(CVPR  = {CVPR})

@String(ICIP  = {ICIP})

@inproceedings{tropf2005region,
  title={Region Segmentation for Facial Image Compression},
  author={Tropf, Alexander and Chai, Douglas},
  booktitle={2005 5th International Conference on Information Communications \& Signal Processing},
  pages={1556--1560},
  year={2005},
  organization={IEEE}
}

@article{elad2007low,
  title={Low bit-rate compression of facial images},
  author={Elad, Michael and Goldenberg, Roman and Kimmel, Ron},
  journal={IEEE Transactions on Image Processing},
  volume={16},
  number={9},
  pages={2379--2383},
  year={2007},
  publisher={IEEE}
}

@article{sujatha2016compression,
  title={Compression Based Face Recognition Using DWT and SVM.},
  author={Sujatha, BM and Madiwalar, Chetan Tippanna and Suresh Babu, K and Raja, KB and Venugopal, KR},
  year={2016},
  publisher={SIPIJ}
}

@inproceedings{wang2019scalable,
  title={Scalable facial image compression with deep feature reconstruction},
  author={Wang, Shurun and Wang, Shiqi and Zhang, Xinfeng and Wang, Shanshe and Ma, Siwei and Gao, Wen},
  booktitle={2019 IEEE International Conference on Image Processing (ICIP)},
  pages={2691--2695},
  year={2019},
  organization={IEEE}
}

@article{zhang2024principal,
  title={Principal Component Approximation Network for Image Compression},
  author={Zhang, Shupei and Zhao, Chenqiu and Basu, Anup},
  journal={ACM Transactions on Multimedia Computing, Communications and Applications},
  volume={20},
  number={5},
  pages={1--20},
  year={2024},
  publisher={ACM New York, NY}
}

@article{wu2020general,
  title={General generative model-based image compression method using an optimisation encoder},
  author={Wu, Mengtian and He, Zaixing and Zhao, Xinyue and Zhang, Shuyou},
  journal={IET Image Processing},
  volume={14},
  number={9},
  pages={1750--1758},
  year={2020},
  publisher={Wiley Online Library}
}

@article{yang2023facial,
  title={Facial image compression via neural image manifold compression},
  author={Yang, Wenhan and Huang, Haofeng and Liu, Jiaying and Kot, Alex C},
  journal={IEEE Transactions on Circuits and Systems for Video Technology},
  year={2023},
  publisher={IEEE}
}

@article{mao2023scalable,
  title={Scalable Face Image Coding via StyleGAN Prior: Towards Compression for Human-Machine Collaborative Vision},
  author={Mao, Qi and Wang, Chongyu and Wang, Meng and Wang, Shiqi and Chen, Ruijie and Jin, Libiao and Ma, Siwei},
  journal={IEEE Transactions on Image Processing},
  year={2023},
  publisher={IEEE}
}

@article{yang2024lossy,
  title={Lossy image compression with conditional diffusion models},
  author={Yang, Ruihan and Mandt, Stephan},
  journal={Advances in Neural Information Processing Systems},
  volume={36},
  year={2024}
}

@inproceedings{careil2023towards,
  title={Towards image compression with perfect realism at ultra-low bitrates},
  author={Careil, Marlene and Muckley, Matthew J and Verbeek, Jakob and Lathuili{\`e}re, St{\'e}phane},
  booktitle={The Twelfth International Conference on Learning Representations},
  year={2023}
}

@article{relic2024lossy,
  title={Lossy Image Compression with Foundation Diffusion Models},
  author={Relic, Lucas and Azevedo, Roberto and Gross, Markus and Schroers, Christopher},
  journal={arXiv preprint arXiv:2404.08580},
  year={2024}
}

@article{brock2018large,
  title={Large Scale GAN Training for High Fidelity Natural Image Synthesis},
  author={Brock, Andrew},
  journal={arXiv preprint arXiv:1809.11096},
  year={2018}
}

@article{goodfellow2014generative,
  title={Generative adversarial nets},
  author={Goodfellow, Ian and Pouget-Abadie, Jean and Mirza, Mehdi and Xu, Bing and Warde-Farley, David and Ozair, Sherjil and Courville, Aaron and Bengio, Yoshua},
  journal={Advances in neural information processing systems},
  volume={27},
  year={2014}
}

@inproceedings{rombach2022high,
  title={High-resolution image synthesis with latent diffusion models},
  author={Rombach, Robin and Blattmann, Andreas and Lorenz, Dominik and Esser, Patrick and Ommer, Bj{\"o}rn},
  booktitle={Proceedings of the IEEE/CVF conference on computer vision and pattern recognition},
  pages={10684--10695},
  year={2022}
}

@inproceedings{CelebAMask-HQ,
  title={MaskGAN: Towards Diverse and Interactive Facial Image Manipulation},
  author={Lee, Cheng-Han and Liu, Ziwei and Wu, Lingyun and Luo, Ping},
  booktitle={IEEE Conference on Computer Vision and Pattern Recognition (CVPR)},
  year={2020}
}

@inproceedings{zhang2018unreasonable,
  title={The unreasonable effectiveness of deep features as a perceptual metric},
  author={Zhang, Richard and Isola, Phillip and Efros, Alexei A and Shechtman, Eli and Wang, Oliver},
  booktitle={Proceedings of the IEEE conference on computer vision and pattern recognition},
  pages={586--595},
  year={2018}
}

@article{ding2020image,
  title={Image quality assessment: Unifying structure and texture similarity},
  author={Ding, Keyan and Ma, Kede and Wang, Shiqi and Simoncelli, Eero P},
  journal={IEEE transactions on pattern analysis and machine intelligence},
  volume={44},
  number={5},
  pages={2567--2581},
  year={2020},
  publisher={IEEE}
}

@article{heusel2017gans,
  title={Gans trained by a two time-scale update rule converge to a local nash equilibrium},
  author={Heusel, Martin and Ramsauer, Hubert and Unterthiner, Thomas and Nessler, Bernhard and Hochreiter, Sepp},
  journal={Advances in neural information processing systems},
  volume={30},
  year={2017}
}

@article{binkowski2018demystifying,
  title={Demystifying mmd gans},
  author={Bi{\'n}kowski, Miko{\l}aj and Sutherland, Danica J and Arbel, Michael and Gretton, Arthur},
  journal={arXiv preprint arXiv:1801.01401},
  year={2018}
}

@inproceedings{wang2019adaptive,
  title={Adaptive wing loss for robust face alignment via heatmap regression},
  author={Wang, Xinyao and Bo, Liefeng and Fuxin, Li},
  booktitle={Proceedings of the IEEE/CVF international conference on computer vision},
  pages={6971--6981},
  year={2019}
}

@article{garcia2017review,
  title={A review on deep learning techniques applied to semantic segmentation},
  author={Garcia-Garcia, Alberto and Orts-Escolano, Sergio and Oprea, Sergiu and Villena-Martinez, Victor and Garcia-Rodriguez, Jose},
  journal={arXiv preprint arXiv:1704.06857},
  year={2017}
}

@inproceedings{cheng2020learned,
  title={Learned image compression with discretized gaussian mixture likelihoods and attention modules},
  author={Cheng, Zhengxue and Sun, Heming and Takeuchi, Masaru and Katto, Jiro},
  booktitle={Proceedings of the IEEE/CVF conference on computer vision and pattern recognition},
  pages={7939--7948},
  year={2020}
}

@article{hific,
  title={High-fidelity generative image compression},
  author={Mentzer, Fabian and Toderici, George D and Tschannen, Michael and Agustsson, Eirikur},
  journal={Advances in Neural Information Processing Systems},
  volume={33},
  pages={11913--11924},
  year={2020}
}

@inproceedings{illm,
  title={Improving statistical fidelity for neural image compression with implicit local likelihood models},
  author={Muckley, Matthew J and El-Nouby, Alaaeldin and Ullrich, Karen and J{\'e}gou, Herv{\'e} and Verbeek, Jakob},
  booktitle={International Conference on Machine Learning},
  pages={25426--25443},
  year={2023},
  organization={PMLR}
}

@article{jpeg2000,
  title={An overview of the JPEG 2000 still image compression standard},
  author={Rabbani, Majid and Joshi, Rajan},
  journal={Signal processing: Image communication},
  volume={17},
  number={1},
  pages={3--48},
  year={2002},
  publisher={Elsevier}
}

@misc{balle_variational_2018,
	title = {Variational image compression with a scale hyperprior},
	url = {http://arxiv.org/abs/1802.01436},
	number = {{arXiv}:1802.01436},
	publisher = {{arXiv}},
	author = {Ballé, Johannes and Minnen, David and Singh, Saurabh and Hwang, Sung Jin and Johnston, Nick},
	urldate = {2023-03-31},
	date = {2018-05-01},
	langid = {english},
	eprinttype = {arxiv},
	eprint = {1802.01436 [cs, eess, math]},
}

@misc{balle_end--end_2017,
	title = {End-to-end Optimized Image Compression},
	url = {http://arxiv.org/abs/1611.01704},
	number = {{arXiv}:1611.01704},
	publisher = {{arXiv}},
	author = {Ballé, Johannes and Laparra, Valero and Simoncelli, Eero P.},
	urldate = {2023-03-31},
	date = {2017-03-03},
	langid = {english},
	eprinttype = {arxiv},
	eprint = {1611.01704 [cs, math]},
}

@inproceedings{he_elic_2022,
	location = {New Orleans, {LA}, {USA}},
	title = {{ELIC}: Efficient Learned Image Compression with Unevenly Grouped Space-Channel Contextual Adaptive Coding},
	isbn = {978-1-66546-946-3},
	url = {https://ieeexplore.ieee.org/document/9879846/},
	doi = {10.1109/CVPR52688.2022.00563},
	shorttitle = {{ELIC}},
	eventtitle = {2022 {IEEE}/{CVF} Conference on Computer Vision and Pattern Recognition ({CVPR})},
	pages = {5708--5717},
	booktitle = {2022 {IEEE}/{CVF} Conference on Computer Vision and Pattern Recognition ({CVPR})},
	publisher = {{IEEE}},
	author = {He, Dailan and Yang, Ziming and Peng, Weikun and Ma, Rui and Qin, Hongwei and Wang, Yan},
	urldate = {2023-04-18},
	date = {2022-06},
	langid = {english},
}

@misc{jiang_mlic_2024,
	title = {{MLIC}++: Linear Complexity Multi-Reference Entropy Modeling for Learned Image Compression},
	url = {http://arxiv.org/abs/2307.15421},
	shorttitle = {{MLIC}++},
	number = {{arXiv}:2307.15421},
	publisher = {{arXiv}},
	author = {Jiang, Wei and Yang, Jiayu and Zhai, Yongqi and Gao, Feng and Wang, Ronggang},
	urldate = {2024-05-28},
	date = {2024-02-19},
	langid = {english},
	eprinttype = {arxiv},
	eprint = {2307.15421 [cs, eess]},
}

@inproceedings{wang2022exploring,
    author = {Wang, Jianyi and Chan, Kelvin CK and Loy, Chen Change},
    title = {Exploring CLIP for Assessing the Look and Feel of Images},
    booktitle = {AAAI},
    year = {2023}
}

@article{liao2024facescore,
  title={FaceScore: Benchmarking and Enhancing Face Quality in Human Generation},
  author={Liao, Zhenyi and Xie, Qingsong and Chen, Chen and Lu, Hannan and Deng, Zhijie},
  journal={arXiv preprint arXiv:2406.17100},
  year={2024}
}

@article{xu2024imagereward,
  title={Imagereward: Learning and evaluating human preferences for text-to-image generation},
  author={Xu, Jiazheng and Liu, Xiao and Wu, Yuchen and Tong, Yuxuan and Li, Qinkai and Ding, Ming and Tang, Jie and Dong, Yuxiao},
  journal={Advances in Neural Information Processing Systems},
  volume={36},
  year={2024}
}

@article{pytorch,
  title={Automatic differentiation in pytorch},
  author={Paszke, Adam and Gross, Sam and Chintala, Soumith and Chanan, Gregory and Yang, Edward and DeVito, Zachary and Lin, Zeming and Desmaison, Alban and Antiga, Luca and Lerer, Adam},
  year={2017}
}

@article{kingma2014adam,
  title={Adam: A method for stochastic optimization},
  author={Kingma, Diederik P},
  journal={arXiv preprint arXiv:1412.6980},
  year={2014}
}

@inproceedings{ffhq,
  title={A style-based generator architecture for generative adversarial networks},
  author={Karras, Tero and Laine, Samuli and Aila, Timo},
  booktitle={Proceedings of the IEEE/CVF conference on computer vision and pattern recognition},
  pages={4401--4410},
  year={2019}
}

@article{bjontegaard2008improvements,
  title={Improvements of the BD-PSNR model},
  author={Bjontegaard, Gisle},
  journal={VCEG-AI11},
  year={2008},
  publisher={ITU-T Q. 6/SG16}
}

@article{wang2024instantid,
  title={Instantid: Zero-shot identity-preserving generation in seconds},
  author={Wang, Qixun and Bai, Xu and Wang, Haofan and Qin, Zekui and Chen, Anthony and Li, Huaxia and Tang, Xu and Hu, Yao},
  journal={arXiv preprint arXiv:2401.07519},
  year={2024}
}

@inproceedings{si2024freeu,
  title={Freeu: Free lunch in diffusion u-net},
  author={Si, Chenyang and Huang, Ziqi and Jiang, Yuming and Liu, Ziwei},
  booktitle={Proceedings of the IEEE/CVF Conference on Computer Vision and Pattern Recognition},
  pages={4733--4743},
  year={2024}
}

@article{saharia2022photorealistic,
  title={Photorealistic text-to-image diffusion models with deep language understanding},
  author={Saharia, Chitwan and Chan, William and Saxena, Saurabh and Li, Lala and Whang, Jay and Denton, Emily L and Ghasemipour, Kamyar and Gontijo Lopes, Raphael and Karagol Ayan, Burcu and Salimans, Tim and others},
  journal={Advances in neural information processing systems},
  volume={35},
  pages={36479--36494},
  year={2022}
}

@inproceedings{zhang2023adding,
  title={Adding conditional control to text-to-image diffusion models},
  author={Zhang, Lvmin and Rao, Anyi and Agrawala, Maneesh},
  booktitle={Proceedings of the IEEE/CVF International Conference on Computer Vision},
  pages={3836--3847},
  year={2023}
}

@inproceedings{li2023blip,
  title={Blip-2: Bootstrapping language-image pre-training with frozen image encoders and large language models},
  author={Li, Junnan and Li, Dongxu and Savarese, Silvio and Hoi, Steven},
  booktitle={International conference on machine learning},
  pages={19730--19742},
  year={2023},
  organization={PMLR}
}

@article{Ng2014ADA,
    author = "Ng, Hongwei and Winkler, Stefan",
    title = "A data-driven approach to cleaning large face datasets",
    journal = "2014 IEEE International Conference on Image Processing (ICIP)",
    year = "2014",
    pages = "343-347"
}

@inproceedings{xiadiffpc,
  title={DiffPC: Diffusion-based High Perceptual Fidelity Image Compression with Semantic Refinement},
  author={Xia, Yichong and Zhou, Yimin and Wang, Jinpeng and An, Baoyi and Wang, Haoqian and Wang, Yaowei and Chen, Bin},
  booktitle={The Thirteenth International Conference on Learning Representations}
}

@article{li2024towards,
  title={Towards extreme image compression with latent feature guidance and diffusion prior},
  author={Li, Zhiyuan and Zhou, Yanhui and Wei, Hao and Ge, Chenyang and Jiang, Jingwen},
  journal={IEEE Transactions on Circuits and Systems for Video Technology},
  year={2024},
  publisher={IEEE}
}

@article{theis2022lossy,
  title={Lossy compression with gaussian diffusion},
  author={Theis, Lucas and Salimans, Tim and Hoffman, Matthew D and Mentzer, Fabian},
  journal={arXiv preprint arXiv:2206.08889},
  year={2022}
}

@article{elata2024zero,
  title={Zero-Shot Image Compression with Diffusion-Based Posterior Sampling},
  author={Elata, Noam and Michaeli, Tomer and Elad, Michael},
  year={2024}
}

@inproceedings{vonderfechtlossy,
  title={Lossy Compression with Pretrained Diffusion Models},
  author={Vonderfecht, Jeremy and Liu, Feng},
  booktitle={The Thirteenth International Conference on Learning Representations}
}

@inproceedings{deng2018arcface,
  title={Arcface: Additive angular margin loss for deep face recognition},
  author={Deng, Jiankang and Guo, Jia and Xue, Niannan and Zafeiriou, Stefanos},
  booktitle={Proceedings of the IEEE/CVF conference on computer vision and pattern recognition},
  pages={4690--4699},
  year={2019}
}

@inproceedings{lifrequency,
  title={Frequency-Aware Transformer for Learned Image Compression},
  author={Li, Han and Li, Shaohui and Dai, Wenrui and Li, Chenglin and Zou, Junni and Xiong, Hongkai},
  booktitle={The Twelfth International Conference on Learning Representations},
  year={2024}
}

@inproceedings{qin2025cassic,
  title={Cassic: Towards Content-Adaptive State-Space Models for Learned Image Compression},
  author={Qin, Shiyu and Wang, Jinpeng and Zhou, Yimin and Chen, Bin and Luo, Tianci and An, Baoyi and Dai, Tao and Xia, Shu-Tao and Wang, Yaowei},
  booktitle={Proceedings of the IEEE/CVF International Conference on Computer Vision},
  pages={15727--15736},
  year={2025}
}

@inproceedings{zeng2025mambaic,
  title={MambaIC: State Space Models for High-Performance Learned Image Compression},
  author={Zeng, Fanhu and Tang, Hao and Shao, Yihua and Chen, Siyu and Shao, Ling and Wang, Yan},
  booktitle={Proceedings of the Computer Vision and Pattern Recognition Conference},
  pages={18041--18050},
  year={2025}
}

@inproceedings{wu2025conditional,
  title={Conditional Latent Coding with Learnable Synthesized Reference for Deep Image Compression},
  author={Wu, Siqi and Chen, Yinda and Liu, Dong and He, Zhihai},
  booktitle={Proceedings of the AAAI Conference on Artificial Intelligence},
  volume={39},
  number={12},
  pages={12863--12871},
  year={2025}
}

@inproceedings{jia2024generative,
  title={Generative latent coding for ultra-low bitrate image compression},
  author={Jia, Zhaoyang and Li, Jiahao and Li, Bin and Li, Houqiang and Lu, Yan},
  booktitle={Proceedings of the IEEE/CVF Conference on Computer Vision and Pattern Recognition},
  pages={26088--26098},
  year={2024}
}


\vfill

\end{document}